\pdfoutput=1
\documentclass{article}
\pdfoutput=1

\usepackage[final]{nips_2017}

\usepackage[utf8]{inputenc} 
\usepackage[T1]{fontenc}    
\usepackage{hyperref}       
\usepackage{url}            
\usepackage{booktabs}       
\usepackage{amsfonts}       
\usepackage{nicefrac}       
\usepackage{microtype}      

\def\bydef{:=}                                        
\def\longversion{1}          

\def\B{\mathbb{B}}

\def\R{\mathcal{R}}

\def\H{\mathcal{H}}

\newtheorem{definition}{Definition}
\newtheorem{lemma}{Lemma}
\newtheorem{remark}{Remark}
\newtheorem{theorem}{Theorem}
\newcommand{\halmos}{\rule[-0.4mm]{2.0mm}{3.2mm}}
\newenvironment{proof}{\par\noindent{\bf Proof\ }}{\hfill\halmos\\[2mm]}
                                        
\title{Critical Hyper-Parameters: No Random, No Cry}

\expandafter\def\expandafter\normalsize\expandafter{%
	    \normalsize
	        \setlength\abovedisplayskip{4pt}
		    \setlength\belowdisplayskip{4pt}
		        \setlength\abovedisplayshortskip{4pt}
			    \setlength\belowdisplayshortskip{4pt}
			    }
\usepackage{tikz}
\def\checkmark{\tikz\fill[scale=0.4](0,.35) -- (.25,0) -- (1,.7) -- (.25,.15) -- cycle;} 
\usepackage{pifont}
\newcommand{\xmark}{\ding{55}}%

\begin{document}

\newcommand{\sylvain}[1]{{\color{cyan} Sylvain: #1}}
\newcommand{\olivierb}[1]{{\color{red} Olivierb: #1}}
\newcommand{\oliviert}[1]{{\color{blue} Oliviert: #1}}
\newcommand{\karol}[1]{{\color{green} Karol: #1}}
\newcommand{\damien}[1]{{\color{tutu} Damien: #1}}
\def\longversion{0}
\def\smallifshort{\ifthenelse{\longversion=0}{\small}{}}
\def\supersmallifshort{\ifthenelse{\longversion=0}{\tiny}{}}

\maketitle
\author{}

{\center{
Olivier Bousquet,
Sylvain Gelly,
Karol Kurach,\\
Olivier Teytaud,
Damien Vincent,
Google Brain, Z\"urich\\
}}
\begin{abstract}
The selection of hyper-parameters is critical in Deep Learning. Because of the
long training time of complex models and the availability of compute resources
in the cloud, ``one-shot'' optimization schemes -- where the sets of
hyper-parameters are selected in advance (e.g. on a grid or in a random manner)
and the training is executed in parallel -- are commonly used.
\cite{BergtraBengio} show that grid search is sub-optimal, especially when
only a few critical parameters matter, and suggest to use random search instead.
Yet, random search can be ``unlucky'' and produce sets of values that
leave some part of the domain unexplored.
Quasi-random methods, such as Low Discrepancy Sequences (LDS)
avoid these issues.
We show that such methods have theoretical properties that make them
appealing for performing hyperparameter search, and demonstrate that,
when applied to the selection of hyperparameters of complex Deep Learning
models (such as state-of-the-art LSTM language models and image classification models), they yield suitable hyperparameters values with much fewer runs than random search.
We propose a particularly simple LDS method which can be used as a
drop-in replacement for grid/random search in any Deep Learning pipeline, both as
a fully one-shot hyperparameter search or as an initializer in iterative batch optimization.
\end{abstract}

\section{Introduction}
Hyperparameter (HP) optimization can be interpreted as the black-box search of an $x$,
such that, for a given function $f:S\subset \R^d\to\R$, the value $f(x)$ is
small, where the function $f$ can be stochastic. This captures the situation
where one is looking for the best setting of the hyper-parameters of a
(possibly randomized) machine learning
algorithm by trying several values of these parameters and picking the value
yielding the best validation error.
Non-linear optimization in general is a well-developed area \cite{nlo}. However,
hyper-parameter optimization in the context of Deep Learning has several
specific features that need to be taken into account to develop appropriate
optimization techniques: function evaluation is very expensive,
computations can be parallelized, derivatives are not easily accessible, and
there is a discrepancy between training, validation and test errors.
Also, there can be very deep and
narrow optima.  Published methods for hyper-parameters search include evolution
strategies \cite{beyer}, Gaussian processes \cite{ego, iago}, pattern
search \cite{toint}, grid sampling and random sampling \cite{BergtraBengio}.
In this work we explore one-shot optimization, hence focusing
on non-iterative methods like random search\footnote{Note that most of the
improvements on one-shot optimization can be applied to batch iterative methods as
well, as shown by Section \ref{iter}.}. This type of optimization is popular - extremely scalable
and easy to implement. Among one-shot optimization methods, grid sampling is
sub-optimal if only a few critical parameters matter because
the same values of these parameters will be explored many times.
Another very popular approach,
random sampling, can suffer from unlucky draws, and leave some part of the
space unexplored. To address those issues, we investigate Low Discrepancy
Sequences (LDS) to see if they can improve upon currently used methods, an approach
also suggested by \cite{BergtraBengio} as future work (although they already proposed the use of
LDS and did preliminary experiments on artificial objective functions, they did not run experiments
on HP tuning for Deep Learning as we do here).
LDS is a rich family of different methods to produce sequences well spread in $[0,1]^d$.
Based on a theoretical analysis and empirical evaluation, we suggest Randomly
Scrambled Hammersley with random shift as a robust one-shot optimization method
for HP search in Deep Learning.

\section{One-shot optimization: formalization \& algorithms}
In this work, we focus on so-called one-shot optimization methods where
one chooses a priori the points at which the function will be evaluated and
does not alter this choice based on observed values of the function.
This reflects the parallel tuning of HPs where one runs the training
of a learning algorithm with several choices of HPs and simply
picks the best choice (based on the validation error).
We do not consider the possibility of saving computational resources by
stopping some runs early based on an estimate of their final performance,
as is done by methods such as Sequential Halving \cite{seqhal} and
Hyperband \cite{hyperband} or validation curves modeling as in \cite{domhan}. However, the techniques presented here can
easily be combined with such stopping methods.

\subsection{Formalization}
We consider a function $f$ defined on $[0,1]^d$ and are interested in finding its infimum.
To simplify the analysis, we will assume that the infimum of $f$ is reached at some point $x^*\in[0,1]^d$.
A one-shot optimization algorithm is a (possibly randomized) algorithm that produces
a sequence $x_1,\ldots,x_n$ of elements of $[0,1]^d$ and, given any function $f$, returns
the value $\min_i f(x_i)$. Since the choice of the sequence is independent of
the function $f$ and its values, we can simply see such an algorithm as producing a
distribution $P$ over sets $\{x_1,\ldots,x_n\}$.
The performance of such an algorithm is then measured via the {\em optimization error}:
$
|\min_i f(x_i)-\inf_{x\in[0,1]^d} f(x)|= |\min_i f(x_i)-f(x^*)|\,.
$
This is a random variable and we are thus interested in its quantiles.
We would like algorithms that make this quantity small with high probability.
Since we have to choose the sequence before having any information about the function,
it is natural to try to have a good coverage of the domain. In particular, by ensuring
that any point of the domain has a close neighbor in the sequence, we can get a control
on the optimization error (provided we have some knowledge of how the function
behaves with respect to the distance between points).
In order to formalize this, we first introduce several notions of how ``spread'' a particular sequence
is.

\begin{definition}[Volume Dispersion]
 The {\em volume dispersion} of $S=\{x_1,\dots,x_n\}\subset [0,1]^d$ with respect to a family
$\R$ of subsets of $[0,1]^d$ is 
$ vdisp(S,\R)\bydef \sup \{\mu(R): R\in\R,\, R\cap S=\emptyset\}\,.  $
\end{definition}
In particular, we will be interested in the case where the family $\R$ is the set $\B$ of all balls 
$B(x,\epsilon)$, in which case instead of considering the volume we can consider directly the radius of the ball.
\begin{definition}[Dispersion]
 The {\em dispersion} of a set $S=\{x_1,\dots,x_n\}\subset [0,1]^d$ is defined as
$ disp(S)\bydef \sup \{\epsilon: x\in[0,1]^d,\, B(x,\epsilon)\cap S=\emptyset\}\,$.
\end{definition}
Since we are interested in sequences that are stochastic, we introduce a more general notion of dispersion.
Imagine that the sets $S$ are generated by sampling from a distribution $P$.
\begin{definition}[Stochastic Dispersion]
The {\em stochastic dispersion} of a distribution $P$ over sets $S=\{x_1,\dots,x_n\}\subset [0,1]^d$
at confidence $\delta \in[0,1]$ is defined as
\begin{eqnarray}
 sdisp(P,\delta) \bydef \sup_{x\in[0,1]^d} \sup \{\epsilon: P(\min_i \|x_i-x\|>\epsilon)\ge 1-\delta \},\\
 sdisp'(P,\delta)\bydef \sup_{x\in[0,1]^d} \sup \{\epsilon: P(\min_i \|x_i-x\|'>\epsilon)\ge 1-\delta \}\,.
\end{eqnarray}
\end{definition}
where $\|.\|'$ is the torus distance in $[0,1]^d$.
We present a simple result that connects the (stochastic) dispersion to the optimization error
when the function is well behaved around its infimum.
\begin{lemma}
Let $\omega(f,x^*\delta)$ be the modulus of continuity of $f$ around $x^*$,
$\omega(f,x^*,\delta)= \sup_{y:\|x^*-y\|\le \delta}|f(x^*)-f(y)|$.
Then for any fixed sequence $S$,
$
 |\min_{x\in S}f(x) - f(x^*)| \le \omega(f, x^*, disp(S))\,,
$
and for any distribution $P$ over sequences,
with probability at least $1-\delta$ (when $S$ is sampled according to $P$),
$
 |\min_{x\in S}f(x) - f(x^*)|\le \omega(f, x^*, sdisp(P, \delta))\,.
$
\end{lemma}
In particular, if the function is known to be Lipschitz then
its modulus of continuity is a linear function $\omega(f, x^*, \delta) \le L(f) \delta$.
In view of the above lemma, the (stochastic) dispersion gives a direct control
on the optimization error (provided one has some knowledge about the behavior
of the function). We will thus present our results in terms of
the stochastic dispersion for various algorithms.

\subsection{Sampling Algorithms}

\subsubsection{Grid and Random}
Grid and random are the two most widely used algorithms to choose HPs in Deep Learning.
Let us assume that $n=k^d$. Then {\bf Grid} sampling consists in
choosing $k$ values for each axis and then taking all $k^d$ points which can be
obtained with these $k$ values per axis.
The value $k$ does not have to be the same for all axes; this has no impact on the present discussion.
There are various tools for choosing the $k$ values per axis; evenly spaced, or purely randomly, or in a stratified manner.
{\bf Random} sampling consists in picking $n$ independently and uniformly sampled vectors in $[0,1]^d$.

\subsubsection{Low Discrepancy Sequences} \label{discdef}

Low-discrepancy sequences have been heavily used in numerical integration but seldom in one-shot optimization.
In the case of numerical integration there exists a tight connection between the integration error
and a measure of ``spread'' of the points called discrepancy.
\begin{definition}[Discrepancy]
The {\em discrepancy} of a set $S=\{x_1,\dots,x_n\}\subset [0,1]^d$ with respect to a family
$\R$ of subsets of $[0,1]^d$ is defined as
$
 disc(S,\R)\bydef \sup_{R\in\R} |\mu_S(R)-\mu(R)|\,,
$
where $\mu(R)$ is the Lebesgue measure of $R$ and $\mu_S(R)$ is the fraction of elements of $S$ that belong to
$R$.
\end{definition}
LDS refer to algorithms constructing $S$ of size $n$ such that  $disc(S, \H_0)=O(log(n)^d/n)$.
One common way of generating such sequences is as follows: pick $d$ coprime integers $q_1,\ldots,q_d$.
For an integer $k$, if $k=\sum_{j\ge 0} b_j q^j$ is its $q$-ary representation, then we define
$\gamma_q(k)\bydef \sum_{j\ge 0}b_jq^{-j-1}$ which corresponds to the point in $[0,1]$ whose
$q$-ary representation is the reverse of that of $k$.
Then the {\bf Halton} sequence \cite{Halton} is defined as $\{(\gamma_{q_1}(k),\ldots,\gamma_{q_d}(k)): 1\le k\le n\}$
and the {\bf Hammersley} sequence is defined as $\{((k-\frac12)/n,\gamma_{q_1}(k),\ldots,\gamma_{q_{d-1}}(k)): 1\le k\le n\}$
One can also define scrambled versions of those sequences by randomly permuting the
digits in the $q$-ary expansion (with a fixed permutation).
The {\bf Sobol} sequence \cite{keyPropertiesSobol}
is also a popular choice and can be constructed using Gray codes \cite{graySobol}. It has
the property that (for $n=2^d$) all hypercubes obtained by splitting each axis into two equal parts
contain exactly one point. We use the publicly available implementation of Sobol by John Burkardt (2009).

It is often desirable to randomize LDS. This improves their
robustness and permits repeated distinct runs.
Some LDS are randomized by nature, e.g. when the scrambling is randomized, as in
Halton scrambling.
Another randomization consists in discarding the first $k$ points, with $k$ randomly chosen.
A simple and generic randomization technique, called {\bf Random Shifting},
consists in shifting by a random
vector in the unit box; i.e. with $x=(x_1,\dots,x_n)$ a sampling in $[0,1]^d$, we
randomly draw $a$ uniformly in $[0,1]^d$, and the randomly shifted counterpart
of $x$ is $mod(x_1+a, 1), mod(x_2+a, 1), \dots, mod(x_n+a,1)$ (where $mod$ is the coordinate-wise
modulo operator).
Random shifting does not change the discrepancy (asymptotically) and 
provides low variance numerical integration \cite{tuffin1996}. The shifted versions of scrambled Halton
and scrambled Hammersley are denoted by S-Ha and S-SH respectively.

\subsubsection{Latin Hypercube Sampling (LHS)}
LHS \cite{lhscitation1,lhscitation2} construct a sequence as follows:
for a given $n$, consider the partition of $[0,1]^d$ into a regular grid of $n^d$ 
cells and then define the index of a cell as its position in the corresponding $n^d$ grid;
choose $d$ random permutations $\sigma_1,\ldots,\sigma_d$ 
of $\{1,\ldots,n\}$ and choose for each $k=1,\ldots,n$, the cell whose index is $\sigma_1(k),\ldots,\sigma_d(k)$,
and choose $x_k$ uniformly at random in this cell.
LHS ensures that all marginal projections on one axis have one point in each of the $n$
regular intervals partitioning $[0,1]$. On the other hand, LHS can be unlucky;
if $\sigma_{j}(i)=i$ for each  $j\in \{1,2,\dots,d\}$ and $i\in \{ 1,2,\dots,n \}$,
then all points will be on the diagonal cells.
Several variants of LHS exist that prevent such issues, in particular
orthogonal sampling \cite{orthogonalSampling}.

\section{Theoretical Analysis}\label{section:theoretical}\label{theory}
Due to length constraints all proofs are reported to the supplementary material.

{\bf{Desirable properties.}}\label{sec:des}
Here are some properties that are particularly relevant in the setup of parallel HP optimization:
 {\bf No bad sequence}: When using randomized algorithms to generate sequences, one desirable property
is that the probability of getting a bad sequence (and thus missing the optimal
setting of the HPs by a large amount) should be as low as possible.
In other words, one would want to have some way to avoid being unlucky.
 {\bf Robustness w.r.t. irrelevant parameters}: 
the performance of the search procedure should not be affected by the addition
of irrelevant parameters. Indeed, it is often the case that one develops
algorithms with many HPs, some of which do not affect the performance
of the algorithm, or affect it only mildly. Otherwise stated, when projecting to a subset
of critical variables, we get a point set with similar properties on this lower dimensional subspace.
As argued in \cite{BergtraBengio}, this concept is critical in HP optimization.
 {\bf Consistency}: The produced sequence should be such that the optimization error
converges to zero as $n$ goes to $\infty$.
 {\bf Optimal Dispersion}:
it is known that the best possible dispersion for a sequence of
$n$ points is of order $1/n^{1/d}$, so it is desirable to have a sequence
whose dispersion is within a constant factor of this optimal rate.
We first mention a result from \cite{rotetichy} which gives an estimate (tight up to constant factors)
of the dispersion (with respect to hyperrectangles) of some known LDS.
\begin{theorem}[\cite{rotetichy}]\label{th:rote}
Consider $S_n$ a set of $n$ Halton points (resp. $n$ Hammersley points, resp. $n$ $(t,m,s)$-net points)
in dimension $d$, then
$
vdisp(S_n, \H) = \Theta(1/n)
$
\end{theorem}
Let $V_d$ be the volume of the unit ball in dimension $d$, $V'_d$ the volume of the orthant of this ball intersecting $[0,\infty)^d$ and $K_d$ be the volume of
the largest hypercube included in this orthant.
The following lemma gives relations between the different measures of spread we have
introduced, showing that the discrepancy upper bounds all the others.
\begin{lemma}[Relations between measures]\label{le:rel}
For any distribution $P$ and any $\delta\ge 0$,
$sdisp(P,\delta)\le sdisp(P, 0)$.
And if $P$ generates only one sequence $S$, then
 $sdisp(P,0)=disp(S)\leq(vdisp(S,\B)/V'_d)^{1/d} \le (K_d vdisp(S,\H)/V'_d)^{1/d}$.
Also we have for any family $\R$, $vdisp(S,\R)\le disc(S,\R)$.
\end{lemma}

\subsection{Dispersion of Sampling Algorithms \& Projection to Critical Variables}
The first observation is that if we compare (shifted) Grid and (shifted) LDS such as considered in Theorem \ref{th:rote} their stochastic dispersion is of the same asymptotic order $1/n^{1/d}$.
\begin{theorem}[Asymptotic Rate]\label{th:asymp}
Consider Random, (shifted) Grid, or a shifted LDS such as those considered in Theorem \ref{th:rote}.
For any fixed $\delta$, there exists a constant $c_\delta$
such that for $n$ large enough,
$
 sdisp(P, \delta) \le K \left(\frac{c_\delta}{n}\right)^{1/d}
 \,.
$
\end{theorem}

\begin{remark}
One natural question to ask is whether the dependence on $\delta$ is different for
different sequences.
We can prove a slightly better bound on the stochastic dispersion for shifted LDS or shifted grid than for random.
In particular we can show that, for all $0<c<\frac12$, for $\delta$ close enough to $1$,
$
 sdisp'(P,\delta) \le K\left(\frac{1-\delta}{n}\right)^{1/d}\,.
$
The constant $K$ depends on the considered sequence and on $\delta$; and for $\delta$
close enough to $1$ the constant is better for S-SH or a randomly shifted grid than for random.
This follows from noticing that when the points of the sequence are at least $2\epsilon$-separated,
	then the probability of $x^*$ to be in the $\epsilon$-neighborhood (for the torus distance) of any point of the sequence
is $n \epsilon^d V_d$.
\end{remark}


Next we observe that the Random sequences have one major drawback which is that they cannot
guarantee a small dispersion, while for Grid or LDS, we can get a small dispersion with
probability $1$. The following theorem follows from the proof of theorem~\ref{th:asymp}:
\begin{theorem}[Guaranteed Success]\label{t}
For Random, $
\lim_{\delta\to 0} sdisp(P,\delta)=1.
$
However, for Grid and Halton/Hammersley, $sdisp(P,0)=O(1/n^{1/d})$.
\end{theorem}
Finally, we show that when the function $f$ depends only on a subset of the variables,
LDS and Random provide lower dispersion while this is not the case for Grid.
Assume $f(x)$ depends only on $(x_i)_{i\in I}$ for some subset $I$ of $\{1,\ldots,d\}$.
Then the optimization error is controlled by the stochastic dispersion of the
projection of the sequence on the coordinates in $I$.
Given a sequence $S$, we define $S_I$ as the sequence of projections on coordinates $I$
of the elements of $S$, and we define $P_I$ as the distribution of $S_I$ when $S$
is distributed according to $P$.
\begin{theorem}[Dependence on Critical Variables]\label{cv}
We have the following bounds on the stochastic dispersion of projections:
	(i) For Random, Halton or Hammersley $sdisp(P_I,\delta)=O(1/n^{1/|I|})$,
	(ii) for Grid $sdisp(P_I,\delta)=\Theta(1/n^{1/d})$.
\end{theorem}

\begin{remark}[Importance of ranking variables properly]
While the above result suggests that the dispersion of the projection only depends on the
number of important variables, the performance in the non-asymptotic regime actually
depends on the order of the important variables among all coordinates.
Indeed, for Halton and Hammersley, the distribution on any given coordinate will become 
uniform only when $n$ is larger than the corresponding $q_i$. Since the $q_i$ have
to be coprime, this means that assuming the $q_i$ are sorted, they will each
be not smaller than the $i$-th prime number. So what will determine the quality of
the sequence when only variables in $I$ matter is the value $\max_{i\in I} q_i$.
Hence it is preferable to assign small $q_i$ to the important variables.
\end{remark}

The conclusion of this section, as illustrated by Fig. \ref{t:checks}
is that among the various algorithms considered here, only the scrambled/shifted
variants of Halton and Hammersley have all the desirable properties from Section \ref{sec:des}.
We will see in Section \ref{endtoend} that, besides having the desirable
theoretical properties, the Halton/Hammersley variants also reach the best
empirical performance.

\subsection{Pathological Functions}\label{sec:path}
As we have seen above, the dispersion can give a characterization of the optimization
error in cases where the modulus of continuity around the optimum is well behaved.
Hence convergence to zero of the dispersion is a sufficient condition for consistency.
However, one can construct pathological functions for which the considered algorithms
fail to give a low optimization error.

{\bf Deterministic sequences are inconsistent}:
In particular, if the sequence is deterministic one can always construct a function on
which the optimization will not converge at all. Indeed, given a sequence $x_1,\ldots,x_n,\,\ldots$,
one can construct a function such that $f(x_i)=1$ for any $i$ and $f(x)=0$ otherwise except in the neighborhood of the $x_i$.
Obviously the optimization error will be $1$ for any $n$, meaning that the optimization procedure
is not even consistent (i.e. fails to converge to the essential infimum of the function 
in the limit of $n\to\infty$).

{\bf Shifted sequences are consistent but can be worse than Random}:
One obvious fix for this issue is to use non-deterministic sequences by adding a random
shift. This will guarantee that the optimization is consistent with probability $1$.
However, since we are shifting by the same vector every point in the deterministic sequence,
we can still construct a pathological function which will be such that the optimization
with the shifted sequence performs significantly worse that with a pure Random strategy.
Indeed, imagine a function which is equal to $1$ on balls $B(x_i,\epsilon)$ (for torus distance) for some small enough $\epsilon$
and which is equal to $0$ everywhere else ($\epsilon$ has to be small enough for the balls not to cover
the whole space, so $\epsilon$ depends on $n$). In this case, the probability that the
shifted sequence gives an optimization error of $1$ is proportional to $\epsilon V_d$,
while it is less than $(n\epsilon V_d)^n$ (which is much smaller for $\epsilon$ small enough) for Random.
This proof does not cover stochastic LDS; but it can be extended to random shifts of any possibly stochastic point set which has a positive probability for at least one fixed
set of values - this covers all usual LDS, stochastic or not.

\begin{figure}
\center
\begin{minipage}{.49\linewidth}{\scriptsize \begin{tabular}{|p{1.4cm}|c|c|c|c|c|}
\hline 
 Property & Grid & Rand & LHS & S-Ha & S-SH \\
\hline  
 No Bad Sequence &  \checkmark  & \xmark & \xmark & \checkmark & \checkmark  \\
\hline 
 Robust Irrelevant Params & \xmark  & \checkmark & \checkmark & \checkmark & \checkmark \\
\hline 
 Consistency & \checkmark  & \checkmark & \checkmark & \checkmark & \checkmark \\
\hline 
 Optimal dispersion & \checkmark & \checkmark & \checkmark & \checkmark & \checkmark \\
\hline 
\end{tabular} 
}
\end{minipage}
\hspace*{-0.2cm}
\begin{minipage}{.51\linewidth}
\includegraphics[width=.9\linewidth]{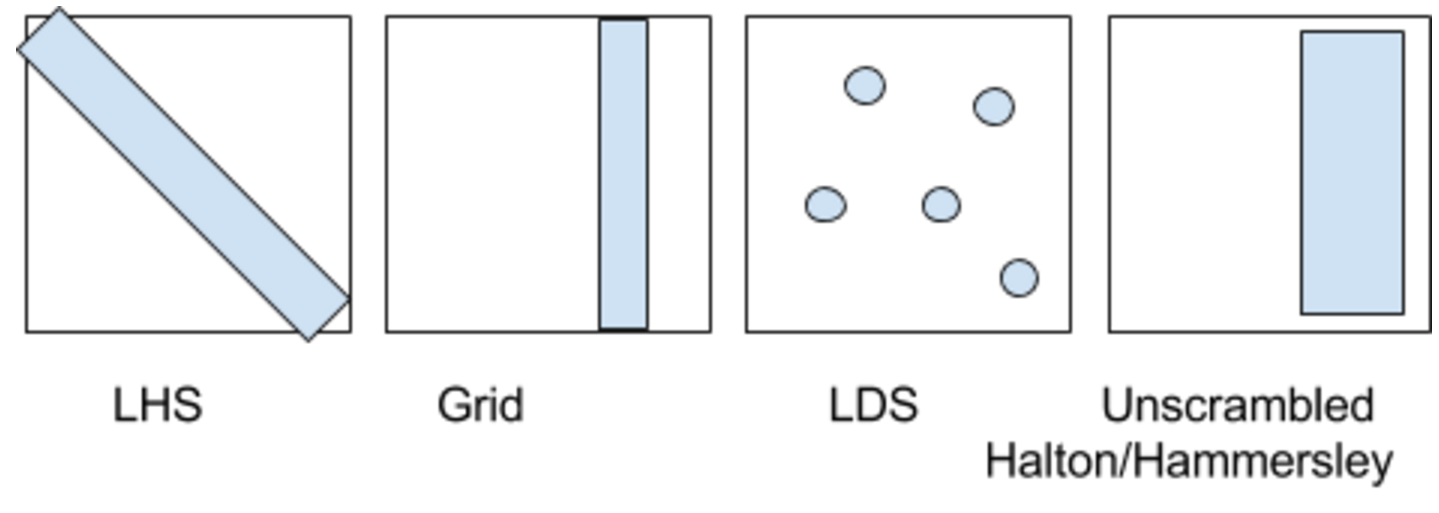}
\scriptsize
\end{minipage}
\caption{Left:\label{t:checks}
Summary of the properties for some of the considered sampling methods.
S-SH and S-Ha have all the desired properties.
  \label{cex}Right:
  Pathological examples where various sampling algorithms will perform
  worse than Random. LHS can produce sequences that are aligned with the
  diagonal of the domain or that are completely off-diagonal with a higher
  probability than Random. This can be exploited by a function with high
  values on the diagonal and low values everywhere else.
  Grid: when the area with low values is thin and depends on one axis only, Grid
  is more likely to fail than Random.
  LDS (with or without random shift): as explained in Section \ref{sec:path},
  if the function values are high around the points in the sequence and low
  otherwise, even with random shift, the performance will be worse than Random.
  Halton or Hammersley without scrambling: due to the sequential nature
  of the function $\gamma_q(k)$, the left part of the domain (lower values)
  is sampled more frequently than the right hand side.
  }
\end{figure}

{\bf Other pathological examples}:
If we compare the various algorithms to Random, it is possible to construct pathological
functions which will lead to worse performance than Random as illustrated on Figure \ref{cex} (right).

\section{Experiments}\label{endtoend}

We perform extensive empirical evaluation of LDS.
We first validated the theory using a set of simple toy optimization problems,
these fast problems provide an extensive validation with negligible p-values.
Results (given in Section \ref{toy}) illustrate each claim - the performance of LDS for one-shot optimization (compared to random and LHS), the positive effect of Hammersley (compared to Halton) and of scrambling,
the additional improvement when variables are ranked by decreasing importance,
and the existence of counter-examples.

We perform some real-world deep learning experiments to further confirm the good properties of S-SH
for hyper-parameter optimization. We finally broaden the use of LDS as a first-step initialization
in the context of Bayesian optimization. Some additional experiments are provided in Section \ref{addit},
the additional results are consistent with the claims.

\subsection{Deep Learning tasks}
\label{sec:eval}

\paragraph{Metrics}
We need to have a consistent and robust way to compare the sampling algorithms
which are inherently stochastic. For this purpose, we can simply measure the \textbf{probability} $p$ that a sampling algorithm $S$ performs better than random search, when both use the same budget $b$ of hyper-parameters trials. From this ``win rate'' probability $p$, we can simply define a \textbf{speed-up} $s$ as $s=\frac{2p-1}{1-p}$. Given two instances of random search $R_1$ and $R_2$, with different budgets $b_1$ and $b_2$, $s$ refers to the additional budget needed for $R_1$ to be better than $R_2$ with probability $p$. That is, if $b_1 = (1+s) b_2$, then $p=\frac{1+s}{2+s}$ or equivalently $s=\frac{2p-1}{1-p}$.

In addition to the speed-up $s$, we report $p$ and, when we have enough experiments on a single setup for having meaningful such statistics, we also report the raw improvements in validation score when using the same budget.


\paragraph{Language Modeling}
We use language modeling as one of the challenging domain. Our datasets include Penn Tree Bank (PTB) \cite{ptb}, using both a word level representation and a byte level representation, and a variant UB-PTB that we created from PTB by randomly permuting blocks of $200$ lines, to avoid systematic bias between train/validation/test\footnote{PTB and Enwik8 have systematic differences between the distributions of the train and the validation/test parts, because it is split in the order of the text. That can create systematic biases when an algorithm is more heavily tuned on the validation set.}. Additionally, we use a subset of the Enwik8~\cite{hutter} dataset, about 6\%, which we name MiniWiki, and the corresponding shuffled version MiniUBWiki.

We use close to state-of-the-art LSTM models as language models, and measure the perplexity (or bit-per-byte) as the target loss. We compare S-SH and LHS with random search on all those datasets, tuning 5 HPs. More details can be found in \textit{setup A} of Section \ref{boringdetails}.

\begin{figure}
\center
\center
\begin{minipage}{.46\linewidth}
\includegraphics[width=1.\linewidth]{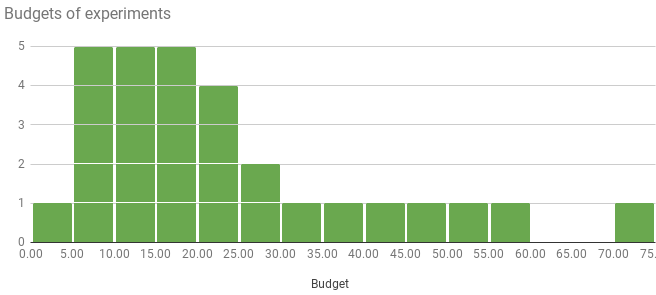}
\end{minipage}
\begin{minipage}{.52\linewidth}
	\scriptsize\begin{tabular}{|@{}c@{}|@{}c@{}|@{}c@{}|@{}c@{}||@{}c@{}|}
\hline
		& budget  & budget$\in$& budget  & All budgets \\
		&  $<11$ & $\{11,\dots, 28\}$ & $> 28$) & \\
\hline
		S-SH win rate & 0.529 $\pm$ 0.04 & {\bf{0.638 $\pm$ 0.03}} & 0.513\% $\pm$ 0.03   &   {\bf{0.569$\pm$0.03}}   \\
		S-SH speed-up     &  +12.3\%           &    {\bf{+76.6\%}}  &   +5.3\%    &   {\bf{+32\%}}  \\
		S-SH bit-per-byte     &        /    &    {\bf{-1.1\%}}  &   /   &   {-0.41\%}  \\
\hline
LHS win rate & {\bf{0.686 $\pm$ 0.06}}  & 0.546 $\pm$ 0.04 & 0.50 $\pm$ 0.07&  0.555$\pm$0.05 \\
		LHS speed-up     & {\bf{ +118\%}} &  +20\%    &  +0\%     & +24.7\%   \\
		LHS bit-per-byte     & / &  -0.08\%    &  /    & {-0.48\%}   \\
\hline
\end{tabular}
\end{minipage}
\caption{Experiments on real-world language modeling, depending on the budget: we provide frequencies at which S-SH (resp. LHS) outperforms random and speedup interpretations.
Left:\label{lhsbudgets} Histogram of budgets used for comparing LHS, S-SH and Random.
Right:\label{zola} Winning rate of S-SH and LHS compared to random, on language modeling tasks (PTB, UBPTB, MiniWiki) with various budgets.}
\end{figure}

The results can be found in Fig. \ref{zola}. Using that setup, both LHS and S-SH outperform random on all considered metrics, and for all budgets. LHS outperforms S-SH for small budgets (<11 values explored), while S-SH is best for larger budgets.

\begin{figure}
	\includegraphics[width=.5\linewidth]{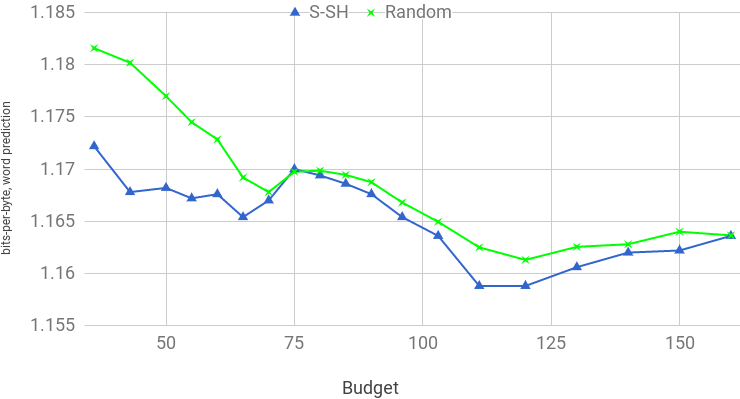}\hspace*{-0.1cm}
  \includegraphics[width=.5\linewidth]{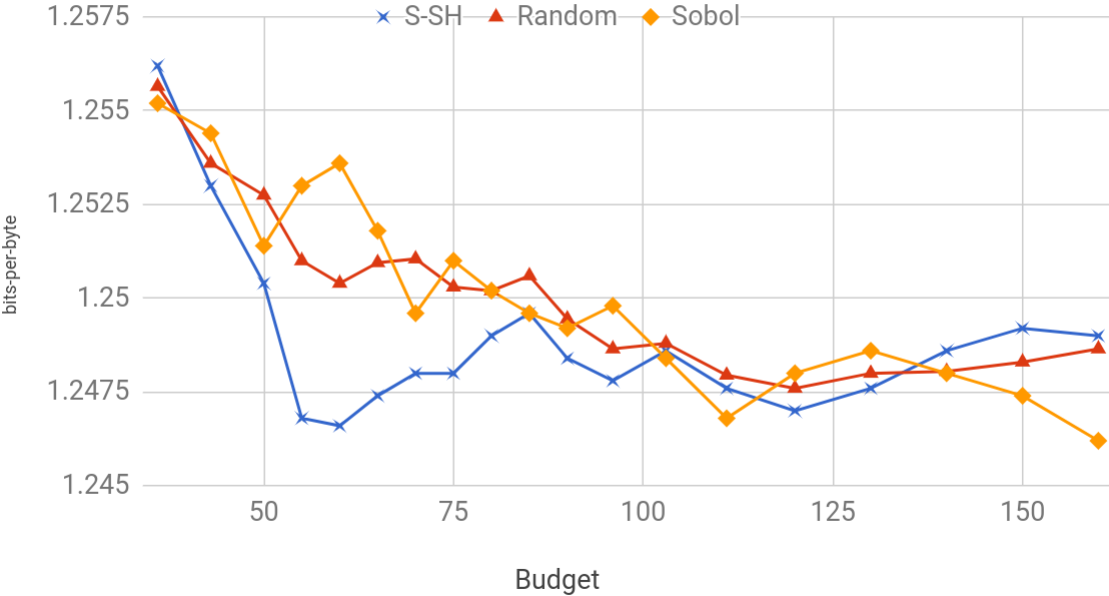}
	\caption{\label{qrptb}
	Comparison between the average loss (bits per byte) for S-SH and for random on PTB-words (left) and PTB-words (right). 
	Moving average (5 successives values) of the performance for setup C in Section \ref{boringdetails} (3 HPs).
  }
\end{figure}

We then perform additional experiments, with larger numbers of epochs and bigger nets, on language modeling.
Results in Fig. \ref{qrptb} confirm that S-SH can provide a significant improvement, especially for budgets 30-60 for the word-PTB model
and for budgets greater than 60 for the byte-PTB model. Detailed p-values for that experiment are given in Section \ref{addit}.

\paragraph{Image classification experiments}
In addition to language modeling, image classification is representative of current Deep Learning challenges. We use a classical CNN model trained on Cifar10~\cite{cifar10}. The chosen architecture is not the state-of-the-art but is representative of the mainstream ones. Those experiments are tuning 6 HPs. More details can be found in \textit{setup B} in Section \ref{boringdetails}.

\begin{figure}
	\center
	\includegraphics[width=.5\linewidth]{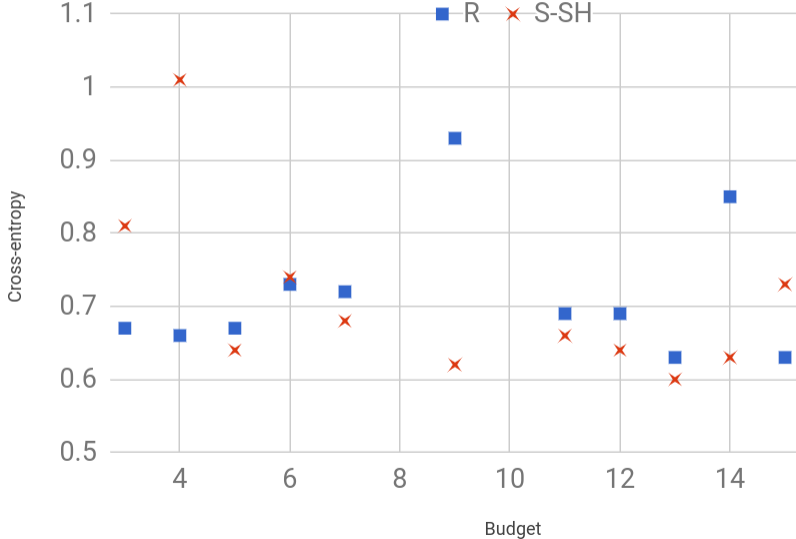}
	\caption{\label{cifarfig}Cifar10: test loss for random and S-SH as a function of the budget.}
\end{figure}

Results are presented in Fig. \ref{cifarfig}. Except for very small budgets ($<5$, for which S-SH fails and for which we might recommend LHS rather than LDS; see also Fig. \ref{zola} on this LHS/S-SH comparison for small budget) we usually get better results with S-SH.


\subsection{LDS as a first-step in a Bayesian optimization framework.}

LDS (here S-Ha) also succeeds in the context of sampling the first batch in a Bayesian optimization run (Table \ref{gptable}), LDS outperforms (a) random sampling, (b) LHS, and (c) methods based on pessimistically fantasizing
the objective value of sampled vectors\cite{fanta}.
\begin{table}
	\setlength{\tabcolsep}{2pt}
	\begin{minipage}{.49\linewidth}
	\center\tiny
	\center Dimension 12 \\
	\begin{tabular}{|c|c|c|c|c|c|c|c|}
		\hline
		 & Rastrigin & Sphere & Ellip. & Styb. & Beale & Branin & 6Hump \\
		\hline
		 \multicolumn{8}{|c|}{1 batch}\\
		\hline
		Pessimistic F. & 1.00 & 1.05 & 1.03 & 1.00 & 0.98 & 1.83 & 1.18 \\
		Random init.   & 1.02 & 1.30 & 1.47 & 1.00 & 1.07 & 1.22 & 1.31 \\
		LHS            & 1.00 & 1.04 & 1.09 & 0.99 & 0.99 & 1.72 & 1.17 \\
		\hline
		 \multicolumn{8}{|c|}{3 batches}\\
		\hline
		Pessimistic F. & 1.01 & 1.06 & 1.26 & 1.00 & 0.83 & 1.20 & 0.91 \\
		Random init.   & 1.02 & 1.21 & 1.20 & 1.00 & 0.84 & 1.08 & 0.90 \\
		LHS            & 1.03 & 0.99 & 1.17 & 1.00 & 0.93 & 1.16 & 0.99 \\
		\hline
		 \multicolumn{8}{|c|}{5 batches}\\
		\hline
		Pessimistic F. & 0.98 & 1.30 & 1.47 & 1.00 & 1.25 & 0.89 & 0.97 \\
		Random init.   & 1.01 & 1.54 & 1.29 & 1.00 & 1.12 & 0.67 & 0.83 \\
		LHS            & 0.98 & 1.44 & 1.17 & 1.00 & 1.34 & 1.01 & 1.03 \\
		\hline
	\end{tabular}
	\newline
	\end{minipage}
	\begin{minipage}{.49\linewidth}
	\center\tiny
	\center Dimension 2 \\
	\begin{tabular}{|c|c|c|c|c|c|c|c|}
		\hline
		 & Rastrigin & Sphere & Ellip. & Styb. & Beale & Branin & 6Hump \\
		\hline
		 \multicolumn{8}{|c|}{1 batch}\\
		\hline
		Pessimistic F. & 0.96 & 1.14 & 0.95 & 0.90 & 0.81 & 0.95 & 1.22 \\
		Random init.   & 1.01 & 1.24 & 1.43 & 0.95 & 1.14 & 0.88 & 1.13 \\
		LHS            & 1.02 & 1.17 & 1.04 & 0.91 & 0.84 & 1.01 & 1.25 \\
		\hline
		 \multicolumn{8}{|c|}{3 batches}\\
		\hline
		Pessimistic F. & 1.12 & 1.38 & 1.24 & 0.03 & 0.82 & 1.47 & .88 \\
		Random init.   & 1.07 & 0.83 & 1.00 & 0.41 & 0.89 & 3.14 & 1.38 \\
		LHS            & 1.01 & 1.45 & 1.48 & 0.06 & 0.73 & 0.78 & 1.19 \\
		\hline
		 \multicolumn{8}{|c|}{5 batches}\\
		\hline
		Pessimistic F. & 0.93 & 1.27 & 1.10 & 0.71 & 1.94 & 1.61 & 0.68 \\
		Random init.   & 1.01 & 1.08 & 1.09 & 0.58 & 5.01 & 1.60 & 1.14 \\
		LHS            & 1.05 & 2.09 & 0.33 & 0.87 & 2.56 & 1.62 & 0.63 \\
		\hline
	\end{tabular}
	\newline
	\end{minipage}
	\newline

	\caption{\label{gptable}Ratio ``loss of a method / loss of LDS-BO method'' (i.e. $>1$ means LDS-BO wins) for 64 runs per batch. 
  In dimension 12 for small number of batches LDS-BO clearly wins against competitors; for larger budgets the pessimistic fantasizing can compete (LDS-BO still usually better); in dimension 2 pessimistic fantasizing and LDS-BO perform best, with a significant advantage compared to LHS-BO and random. 
  Fig. \ref{gptable16} (supplementary material) shows that on the other hand, and consistently with known bounds on discrepancy, LHS-BO outperforms LDS-BO for small budget.}
\end{table}
For rare cases Bayesian optimization based on LHS or pessimistic fantasizing outperform our LDS-based Bayesian optimization even for one single batch - 
but in dimension 12 LDS-BO (LDS-initialized Bayesian optimization) still performed best for most numbers of batches in $\{1,3,5\}$ and most functions in $\{Branin, Beale, SixHump, Rastrigin, Ellipsoidal, Sphere\}$. Results were mixed in dimension 2.


\section{Conclusion}
We analyzed theoretically and experimentally the performance of LDS for hyper-parameter
optimization. The convergence rate results show that LDS are strictly better than
random search both for high confidence and low confidence quantiles (Theorem 3
and Remark 1). In particular, with enough points, LDS can find the optimum within
$\epsilon$ distance with probability exactly 1, due to the absence of unlucky
cases, unlike random search. For intermediate quantiles the theoretical rate
bounds are equivalent between LDS and random search but we could not prove that the constant
is better except in dimension 1. In the common case where parameters have different impacts, and only a few really
matter, we show that parameters should be ranked (Remark 2). Further we prove
that LDS outperforms grid search (in case some variables are critically important)
and the best LDS sequences are robust against
bad parameter ranking (Theorem 4). 
The experiments confirmed and extended our theoretical results: LDS is
consistently outperforming random search, as well as LHS except for small budget.
Importantly, while the theory covers only the one-shot case, we get great performance
for LDS as a first step in Bayesian optimization\cite{ego}, even compared to entropy
search\cite{fanta} or LHS-initialized Bayesian optimization.

As a summary of this study, we suggest Randomly Scrambled Hammersley with random
shift as a robust one-shot optimization method.
Replacing random search with this
sampling method is a very simple change in a Deep Learning training pipeline,
can bring some significant speedup
and
we believe should be adopted by most practitioners - though for budgets $<10$ we might prefer LHS. 
To spread the use of LDS as a way to optimize HPs, we will open-source a small library at \href{https://AnonymizedUrl}{https://AnonymizedUrl}.
Our results also support the application of LDS for the first batch of Bayesian optimization.
Importantly, not all LDS perform equally - we were never disappointed by S-SH, but Sobol (used in \cite{sobolhp})
or unscrambled variants of Halton are risky alternatives.

\newpage
\bibliographystyle{abbrv}
\bibliography{../bigbib.bib}

\appendix
\cleardoublepage
{\Large{\center{ 
Supplementary material --- critical hyperparameters: no random, no cry.
}}}

\section{Proofs}

\begin{proof}[Lemma \ref{le:rel}]
A ball of radius $\epsilon$ has volume $V_d \epsilon^d$, so if $v$ is the
volume of a ball, its radius is $(v/V_d)^{1/d}$.
If $v$ is the largest volume of a ball not containing any point of $S$,
then the largest hypercube contained in this ball, which has
volume $v.K_d$ will also be empty, which shows that $vdisp(S,\B)\le K_d vdisp(S,\H)$.
Finally it is easy to see that the volume dispersion is upper bounded
by the discrepancy as it follows from their definition.
\end{proof}

\begin{proof}[Theorem \ref{th:asymp}]
For Grid this is a known result\cite{sukharev}. For Random, the probability
of $n$ randomly picked balls of radius $\epsilon$ to contain any particular point $x^*$ is upper
bounded by $1-(1-V'_d\epsilon^d)^n$, hence this gives an upper bound of
$\left(\frac{\log 1/\delta}{n}\right)^{1/d}$ for $n$ large with respect to $\log 1/\delta$.

For the LDS sequences, this follows from combining Lemma \ref{le:rel} with Theorem \ref{th:rote}.
\end{proof}

\begin{proof}[Theorem \ref{cv}]
The projection of Random (resp. Halton) sequences are Random (resp. Halton) sequences.
For Hammersley the first coordinate is uniform, so the projection onto $I$ is Hammersley
if $1\in I$ or is Halton otherwise.
Theorem \ref{th:asymp} allows to conclude. The result for Grid is obvious.
\end{proof}

\section{Detailed experiment setups}\label{boringdetails}
\paragraph{Setup A} Language model with 3 layers of LSTM with 250 units trained for 6 epochs. In this setup, we have 5 hyper-parameters: weight init scale $[0.02,1]$, Adam's $\epsilon$ parameter $[0.001,2]$, clipping gradient norm $[0.02,1]$, learning rate $[0.5,30]$, dropout keep probability $[0.6,1]$.

\paragraph{Setup B} Image classification model with 3 convolutional layers, with filter size 7x7, max-pooling, stride 1, 512 chanels; then one convolutional layer with filter size 5x5 and 64 chanels; then two fully connected layers with 384 and 192 units respectively. We apply stochastic gradient descent with batch size 64, weight init scale in $[1,30]$ for feedforward layers and $[0.005, 0.03]$ for convolutional layers, learning rate $[0.00018,0.002]$,  epoch for starting the learning rate decay $[2,350]$,  learning rate decay in $[0.5, 0.99999]$ (sampled logarithmically around $1$), dropout keep probability in $[0.995,1.]$. Trained for 30 epochs.

\paragraph{Setup C} Language model with 2 stacked LSTM with 650 units, with the following hyper-parameters: learning rate $[0.5,30]$, gradient clipping $[0.02,1]$, dropout keep probability $[0.6,1]$. Model trained for 45 epochs.

\section{Toy optimization problems}\label{toy} 
We conducted a set of experiments on multiple toy problems to quickly validate our assumptions.
Compared methods are one-shot optimization algorithms based on the following samplings:
Random, LHS, Sobol, Hammersley, Halton, S-SH.
We loop over dimensions 2, 4, 8, and 16; we check three objective functions (l2-norm $f(x)=||x-x^*||$, illcond 
 $f(x) = \sum_{i=1}^d (d-i)^3(x_i-x^*_i)^2$, reverseIllcond $f(x)= \sum_{i=1}^d (1+i)^3(x_i-x^*_i)^2$). 
 The budget is $n=37$ in all cases.
We used antithetic variables, thanks to mirroring w.r.t the 3 first axes (hence
8 symmetries). Each method is tested with and without this 3D mirroring. 3D mirroring
deals conveniently with multiples of 8; additional points are generated in a pure
random manner. $x^*$ is randomly drawn uniformly in the domain. Each of these 12
experiments is reproduced 1221 times.

In each of these 12 cases (4 different dimensions $\times$ 3 different test functions), on average over the 1221 runs, Sobol and
      Scrambled-Hammersley performed better than Random in terms of simple regret. 
      This validates, on these artificial problems,
      both Sobol and Scrambled-Hammersley, in terms of one-shot optimization and 
      face to random search, with {\bf{p-value 0.0002}}.

Unsurprisingly, for illcond, Halton and Hammersley outperform random, whereas
      it is the opposite for reverseIllcond at least in dimension 8 and 16,
      i.e. it matters to have the most important variables first for these sequences. Scrambled versions (both Halton and Hammersley) resist
      much better and still outperform random - this validates scrambling.

Scrambled Hammersley performs best 6 times, scrambled Halton and
      Hammersley 2 times each, Sobol and Halton once each;  none of the 3d
      mirrored tools ever performed best. This invalidates mirroring, and confirm the
      good behavior of scrambled Hammersley.

These intensive toy experiments are in accordance with the known results and our
theoretical analysis (Section~\ref{section:theoretical}). Based on the results,
the randomly shifted Scrambled-Hammersley is our main LDS for the evaluation on
real Deep Learning tasks.

\section{Artificial datasets}\label{appendix:toydata}

Our artificial datasets are indexed with one string (AN or ANBN or .N or others, detailed below) and 4 numbers.
For the artificial dataset $C$, the 4 numbers are the vocabulary size (the number of letters),
maximum word size ($n$ or $N$), vocabulary growth and depth - unless specified otherwise, vocabulary growth is 10
and depth is 3.
	 There are also four parameters for ANBN, AN, .N and anbn; but the two last parameters are different: they are ``fixed size'' (True for fixed size, False otherwise) and ``size gaps'' (impact of size gaps equal to True detailed below). For example, artificial(anbn,26,10,0,1) means that n is randomly drawn between 1 and 10, and that there are size gaps; whereas artificial(anbn,26,10,1,0) means that n is fixed equal to 10 and there is no size gap. 
\begin{itemize}
	\item AN is a artificial dataset with, as word, a single letter randomly drawn (once for each sequence)
  and repeated a fixed number of times (same number of times for different sequences, but different letter).
  For examples, the first sequence might be ``qqqqqq qqqqqq qqqqqq'' and the second one ``pppppp pppppp pppppp''.

	\item The ``ANBN'' testbed is made of words built by concatenating N copies of a given randomly drawn letter, followed by N copies of another randomly drawn letter.
The words are repeated until the end of the sequence. For different sequences, we have different letters, but the same number N.
	\item In the ``ABNA'' dataset, a word is one letter (randomly drawn, termed A), then N copies of another letter (randomly drawn and termed B),
  and then the first letter again.

	\item We also use the ``.N'' testbed, where the ``language'' to be modelled is made of sequences, each of them containing only one word (made of $N$ randomly independently drawn letters) repeated until the end
of the sequence. The first sequence might be ``bridereix bridereix bridereix'' and the second sequence ``dunlepale dunlepale dunlepale''.
	\item We also have ``anbn'' as a testbed: compared to ANBN, the number of letters vary for each word in a same sequence, and the letters vary even inside a sequence.
  The first sequence might be ``aaabbb ddddcccc db''.
  
	\item Finally, we use the ``C'' testbed, in which there are typically 26 letters (i.e. the vocabulary size is 26 unless stated otherwise), words are randomly drawn combinations of letters and there 
exists $V\times26$ words of e.g. 10 (word size) letters; and there exists $V^2\times 26$ groups (we might say ``sentences'') of 10 words, where $V$ is the ``vocabulary growth'';
there are 3 levels (letters, words, groups of words) when the depth is $3$. For example $artificial(C,26,10,7,3)$ contains $26$ letters, 26$\times$7 words of length $10$,
and 26$\times 7^2$ groups of $10$ words. 

\end{itemize}

For each sequence of these artificial datasets, the word size N is randomly chosen uniformly between 1 and the maximum word size, except when fixed size is 1 - in which case the word size is always the maximum word size.

If size gaps is equal to True, then the test sizes used in test and validation
are guaranteed to not have been seen in training; 4 word sizes are randomly chosen for valid and for test, and the other 16 are used in training. There are
10 000 training sequences, 1000 validation sequences, 1000 test sequences. Each sequence is made of 50 words, except for C for which a sequence is one group of the maximum level.

In all artificial sequences, letters which are not predictable given the type of sequence have a weight 0 (i.e. are not taken into account when computing the loss).
In all cases, the loss functions are in bits-per-byte.


\section{Additional experiments}\label{addit}
\subsection{Robustness speed-up}
In addition to the speed-up defined in \autoref{sec:eval}, we here define a {\bf{robustness speed-up}}.

Speed-up was defined as such because $(1+s)/(2+s)$ is the frequency at which a random search with budget $b\times(1+s)$
wins against a random search with budget $b$. We can define the speed-up similarly when $S$ competes
with several (say $K$) instances of random search with the same budget $b$: $p=(1+s)/(K+1+s)$.
Note that when $K>1$, we could define another type of speed-up: instead of the frequency at which $S$ is {\em{better than all}} $K$ independent copies of random search, we can consider the frequency at which it performs {\em{worse than all}} these $K$ instances. We call it {\bf{robustness speed-up}}; this speed-up is positive when the algorithm is robust and avoids bad cases. When ambiguous, we refer to the initial one as {\bf{optimistic speed-up}}, otherwise simply {\bf{speed-up}}.

\subsection{Average speed-ups over many budgets and testbeds: LDS outperforms random in a stable manner}\label{diversetests}
We consider tuning three commonly used hyper-parameters ranked in the following way:
learning rate, weight init scale, max gradient norm in three different settings -- untuned, half-tuned
and tuned -- which corresponds to some degree of prior on the expected range of those hyper parameters
(Table~\ref{table:rangehps}).
\begin{table}
	\center
\begin{tabular}{|c|c|c|c|}
\hline
  & Untuned & Half-tuned & Tuned \\
\hline
	Learning rate & $[0.001, 3000]$ & $[0.1, 200]$  & $[0.5, 30]$  \\
	Max grad. norm & $[0.002, 100]$ & $[0.006, 30]$ & $[0.02, 1.0]$ \\
	Weight init scale & $[0.02, 1]$ & $[0.02, 1]$ & $[0.02, 1]$ \\
\hline
\end{tabular}
\caption{\label{table:rangehps}Range of hyper-parameters.}
\end{table}

\begin{table}
\scriptsize\centering
\begin{minipage}{.49\linewidth}
	\center
	\begin{tabular}{|@{}c@{}@{}c@{}|@{}c@{}|@{}c@{}|}
\hline
Setting       & Budgets  & Optimistic &  Robustness \\
&             & speed-up &   speed-up     \\
\hline
\hline
Untuned*         & 5:12  & +350\%$^o$ &  \\
(3 HP, 7 epochs) &       &                      & \\
\hline
Untuned*         & 5:15, 17, 19 & +237\%$^o$ &  \\
(3 HP, 7 epochs) &       &                          & \\
\hline
Untuned*   & 25:5:145    & +12.8 \%$^o$ & +26\%    \\   
(3 HP, 7 epochs) &       & (x)                                 &               \\
\hline
Half-tuned*& 40:10:120   & +50.0 \%$^o$ &  +25\%    \\   
(3 HP, 7 epochs)        & and 140:20:200 &                  &                \\
\hline
\end{tabular}
\end{minipage}
\begin{minipage}{.49\linewidth}
	\center
	\begin{tabular}{|@{}c@{}@{}c@{}|@{}c@{}|@{}c@{}|}
\hline
Setting       & Budgets              & Optimistic &  Robustness \\
&                            & speed-up &   speed-up     \\
\hline
\hline
Tuned  & & & \\
(3 HP, 7 epochs)*        & 5:47        & -51\%    &   +28\% \\
\hline
Tuned*         & 48:73   & +87\%   &  +50\% \\
(3 HP, 7 epochs) &                &                   &   \\
\hline
Tuned*           & 20:5:30     & +160\%$^o$  & +160\%  \\
(3 HP, 4 epochs) &             &         & \\
\hline
Leaky**    & 10:5:100   &  +62 \% &  +40\% \\  
(5 HP, 30 epochs)       &           &            &            \\
\hline
PTB Bytes/Words*** & budget 20  & +57\% & +57\% \\   
(7 HP, 17 epochs)  & budget 30  & +0\% & +0\% \\   
(36 xps) & & & \\
\hline
\end{tabular}
\end{minipage}
 \caption{Performance of Shifted Scrambled-Hammersley versus random search in various setups.
	*: PTB bytes and words, Toy sequences (see Section \ref{appendix:toydata}) C, AN, anbn and ABNA, no size gap, no fixed size, 
  3 instances of random vs S-SH, 200 units, 11 epochs. 
	**: leaky setup as defined in Section \ref{leaky}, toy sequences .N, AN, anbn with 26 letters and max word size 5 (resp. 10), 2 Lstm, 100 units, with size gaps, variable word size, 3 instances of random vs S-SH.
	***: learning rate in $[0.5,30]$, max gradient norm $[2e-3,1]$, Adam's epsilon $[0.001,2]$, weight init scale $[2e-3,10]$, number of epochs before learning rate decay $[5,15]$, dropout keep probability $[0.2,1]$, 17 epochs, 2 LSTM with 650 units, S-SH vs one instance of random.
	${}^o$ denote p-values $<0.05$; there are 
	cases in which the speed-up was negative but these cases are not statistically significant; and when the speed-up was less than $+200\%$ we also checked
	the robustness speed-up and it was positive in all cases - this illustrates the robustness property emphasized in Section \ref{theory}.
  (x): Fig. \ref{qrsmallptb} suggests a speed-up $>100\%$ for budget $\simeq 20-70$ decreasing to zero for large budget.}\label{table:main}\label{main_table}
  \end{table}

\begin{figure}
\includegraphics[width=.5\linewidth]{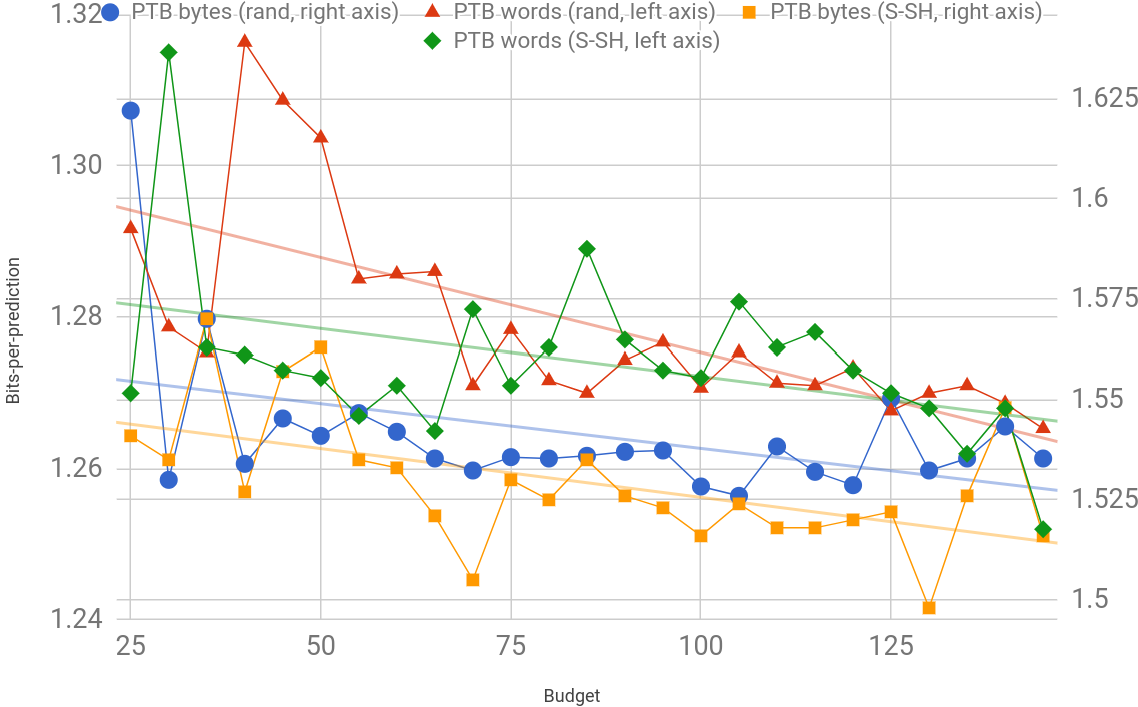}\hspace*{-0.1cm}
\caption{\label{qrsmallptb}Language modeling with moderate networks, (x) in Table \ref{table:main}.}
\end{figure}
S-SH performs clearly better than random (Table \ref{table:main}): all statistically significant results are in favor
of S-SH compared to random. In cases in which the speed-up was negative, this was not statistically significant;
and we checked, in all cases in which the speed-up was less than +200\%, that the robustness speed-up is still positive.


\subsection{Results on toy language modeling problems with additional model hyper-parameters.}\label{leaky}
\begin{figure}
	\center
\includegraphics[width=.75\linewidth]{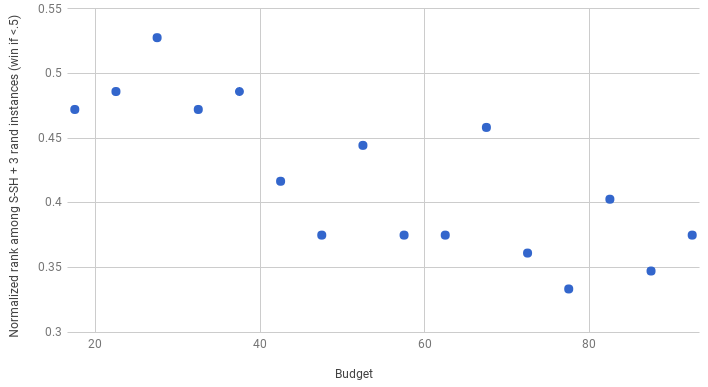}
	\caption{\label{signif_leak}
	Rank of S-SH among S-SH and 3 random instances, normalized to $[0,1]$, on toy language modelling testbeds and with leaky LSTM; $<0.5$ in most cases i.e. success for S-SH. Points are moving averages over 4 independent successive budgets hence points 4 values apart are independent. This corresponds to an overall speed-up of +62\%, i.e. we get with 50 runs the same best performance as random with 81 runs.}
\end{figure}

We use a modified version of the LSTM cell using 100 leaky state units;
each state unit includes 2 additional parameters,
sampled at initialization time according to a gaussian distribution, whose mean and variance
are the 2 additional HPs of the model. Whereas we use, in all experiments (including the present ones), learning rate and gradient clipping norm as the
two first parameters, a posteriori analysis shows that the important parameters are these two specific parameters and therefore our
prior (on the ranking of HP) was wrong - this is a common scenario where we do not have any prior
on the relative importance of the hyper-parameters. The model was trained on $6$ of the artificial datasets described in
Appendix~\ref{appendix:toydata}, namely toy(.N26,5,0,1), toy(.N,26,10,0,1),	toy(AN,26,5,0,1),	toy(AN,26,10,0,1), toy(anbn,26,5,0,1),	toy(anbn,26,10,0,1).
We use learning rate $\in[5,100]$, weight init scale $\in[0.05, 1]$, max gradient norm $\in[0.05, 1]$,
	and the two new parameters (mean/std) are respectively $\in[-9,9]$ and $\in[0.01, 10]$. We check budget $10,15,20,25,\dots,100$.
Fig. \ref{signif_leak} shows
the rank of S-SH, among S-SH and 3 instances of random search; this rank is between 1 and 4; after normalization to $[0,1]$ we get 0.404$\pm$ 0.097, 0.456 $\pm$ 0.084, 0.281$\pm$ 0.084, 0.491$\pm$0.080, 0.544$\pm$ 0.092, 0.316$\pm$ 0.081 on these 6 tasks respectively; all but one are in favor of S-SH (i.e. rank $<0.5$), 2 are statistically significant, and when agregating over all these runs we get an average rank 0.415 at $\geq$ 3 standard deviations from .5, hence clearly significant.

\subsection{Other sampling algorithms: S-SH performs best}

\subsubsection{Validating scrambling/Hammersley on PTB: not all LDS are equivalent}                   
\begin{figure}                   
\center                   
\includegraphics[width=.8\linewidth]{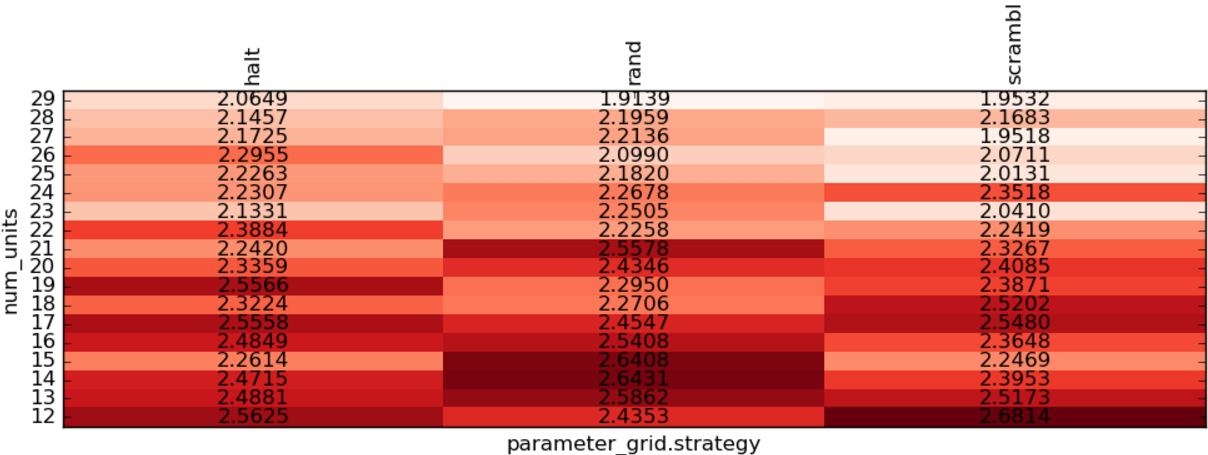}                   
\includegraphics[width=.8\linewidth]{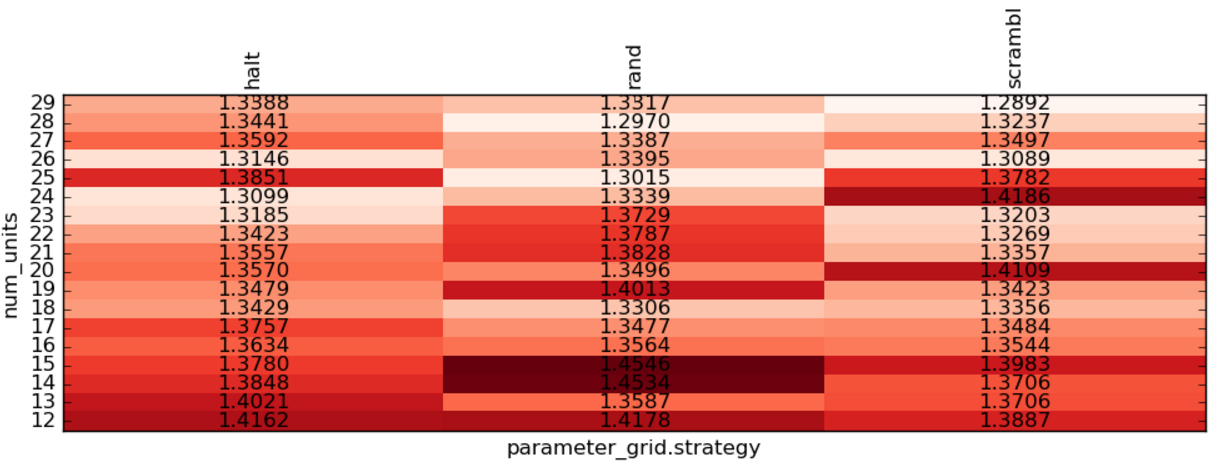}                   
	\caption{\label{scr}Bits-per-byte loss for various settings. Comparing a sophisticated Halton to a naive Halton and to random with 17 learning epochs. Halton refers to the original Halton sequence. Scramb. refers to Hammersley with scrambling - which has the best known discrepancy bounds (more precisely the best proved bounds are obtained by Atanassov's scrambling; for the random scrambling we use, as previously mentioned, it is conjectured that the performance is the same). Top: PTB-Bytes (bits per byte). Bottom: PTB-Words (bits per word). Scrambled-Hammersley outperforms the simple Halton and pure random (see text).}                   
\end{figure}

\begin{figure}                   
\center                   
\includegraphics[width=.8\linewidth]{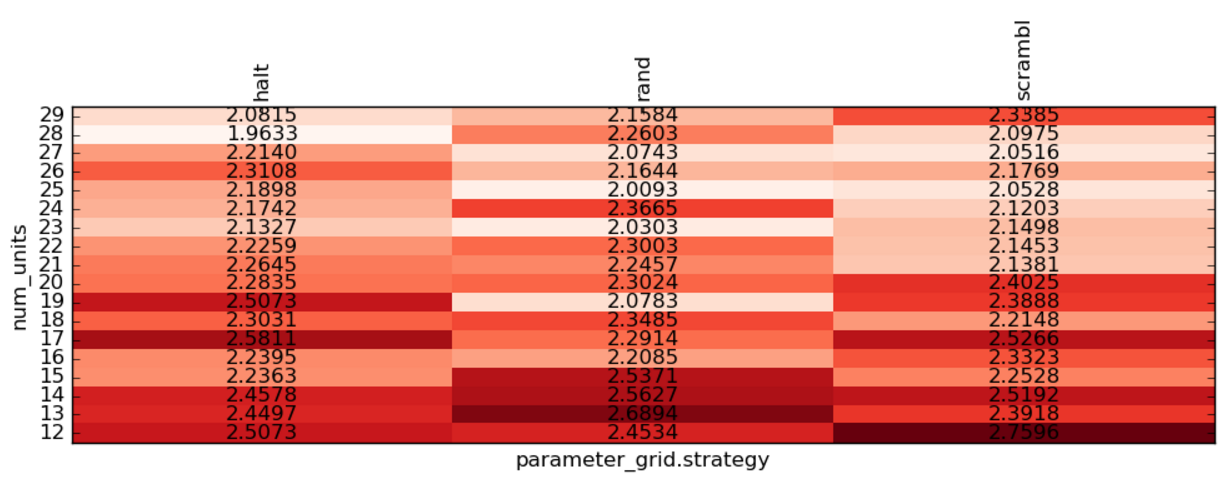}                   
\includegraphics[width=.8\linewidth]{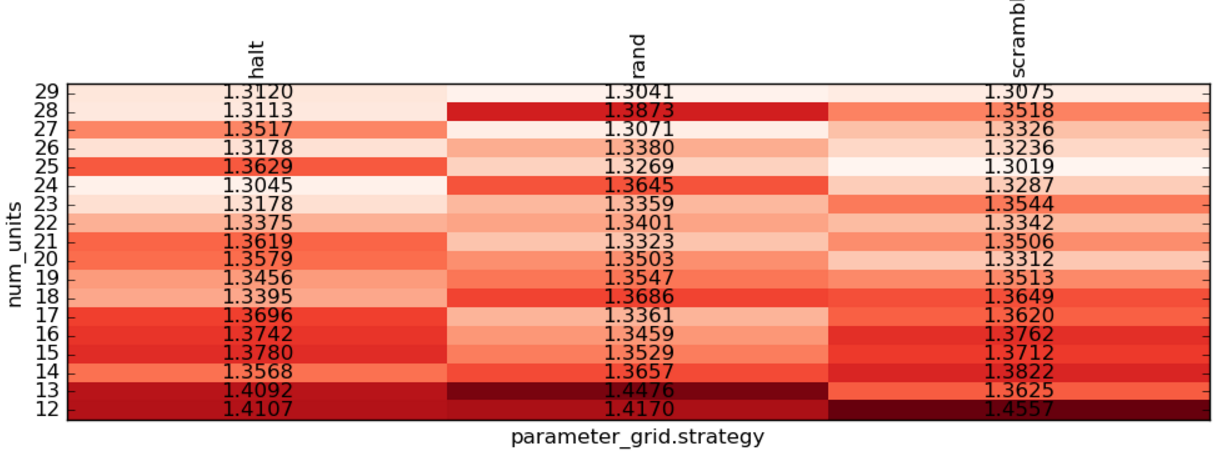}                   
\caption{\label{scrambling2}Same as Fig. \ref{scr} but with budget 30 instead of 20, still 17 learning epochs. Top: PTB-Bytes. Bottom: PTB-Words.                    
No statistically significant difference neither for bytes (9 wins for 18 test cases, 9/18 and 10/18 for S-SH vs random,                   
	for Halton vs rand, and for S-SH vs Halton respectively) or words (9/18, 10/18 and 9/18 respectively).}                   
\end{figure}

Fig. \ref{scr} compares pure random, Halton, and scrambled Halton on PTB-Bytes                   
and PTB-Words. The setting is as follows: 7 HPs (dropout keep probability in $[0.2,1]$,                    
learning rate in $[0.05,300]$, gradient clipping norm in $[0.002,1]$, Adam's epsilon parameter in $[0.001,2]$,                    
weight initialization scale in $[0.002,10]$, epoch index for starting the exponential decay of                    
learning rate in $[5,15]$, learning rate decay in $[0.1,1]$);                   
17 training epochs, 2 stacked LSTM; we perform experiments for a number of units ranging from 12 to 29.                   
The budget for the randomly drawn HPs is 20.                   
Overall, there are 36 comparisons (18 on PTB-Bytes, corresponding to 18 different numbers of units, and 18 on PTB-Words);                    
\begin{itemize}                   
	\item Scrambled-Hammersley outperforms Halton 25 times (p-value {\bf{0.036}});                   
	\item Scrambled-Hammersley outperforms random 22 times (p-value 0.12);                   
\item Halton outperforms random 19 times (p-value 0.43).                   
\end{itemize}                   
The experiment was reproduced a second time, with budget $=$ 30 random tries for the HPs. Results are presented in Fig. \ref{scrambling2} - no statistically significant improvement.                   
\subsection{Robustness vs peak performance}
We have seen in Section \ref{diversetests} that S-SH was always beneficial in terms
of robustness; and most often in terms of peak performance (optimistic speed-up). We develop this point by performing additional experiments. We prefer
challenges, so we focus on the so-called tuned-setting in which our results
in Section \ref{diversetests} were least positive for S-SH. We performed experiments with many different
values of the budget (number of HPs vectors sampled) and tested if S-SH performs better than random
in both low budget regime ($\leq 48$) and large budget $>48$). Each budget from 5 to 73 was tested.
For large budget regime, S-SH performed excellently, both in terms of best performance (speed-up $+87\%$, p-value 0.09)
and worst performance (robustness speed-up $+50\%$, p-value 0.18). 
For budget $\leq 48$, random was performing better than S-SH for the optimistic speed-up ($-51\%$, p-value 0.18); but worse than S-SH for the robustness speed-up ($+28\%$, p-value 0.09).
Importantly, this section focuses on the only setting in which S-SH was not beneficial in our diverse experiments (Section \ref{diversetests});
we have, on purpose, developped precisely the case in which things were going wrong in order to clarify to which extent replacing random by S-SH can be detrimental.

\def\notinthemainpaper{
TODO rewrite this from scratch

We now provide more detailed experimental results for the case in which QR performed poorly,                   
namely the tuned setting, which corresponds to our platform tuned over testbeds used in                   
these experiments. We tested extensively with an increasing sampling number, and report                    
the best/worst in each case; 3 independent random instances are compared to a scrambled                    
Hammersley; which means that under null assumption (the null assumption is that                    
scrambled Hammersley performs equally                    
to random) we would get Scrambled Hammersley best 25\% of runs and worst 25\% of runs;                   
in particular, we would have Scrambled Hammersley the same number of times on both sides.                    
Results are presented in Table \ref{dependency_numtries} and suggest that S-SH                    
\begin{itemize}                   
	\item is outperformed by random for low budget ($\leq 30$) in terms of frequency of best performance ({\bf{p-value 0.026}});                   
	\item but performs better than random for low budget ($\leq 30$) in terms of frequency of worst performance (p-value 0.18);                   
\end{itemize}                   
and then becomes competitive at budget circa 30 and clearly better at                   
budget circa 55; overall, S-SH:                   
\begin{itemize}                   
	\item is outperformed by random for low budget ($< 48$) in terms of frequency of best performance (p-value 0.18, speed-up $-51\%$);                   
	\item but performs better than random for low budget ($< 48$) in terms of frequency of worst performance (p-value 0.09; robustness speed-up $+28\%$);                   
	\item performs better than random for larger budget ($\geq 48$) both in terms of best performance (speed-up $+87\%$) and worst performance (robustness speed-up $+50\%$).     
\end{itemize}                   
\begin{table}                   
\smallifshort                   
\center                   
\begin{tabular}{|c|c|c|}                   
\hline                   
Budget & Worst method & Best method \\                   
\hline                   
\hline                   
5&	     S-SH 	 & random \\                   
6-9&     random 	 & random \\                   
9&	     S-SH 	 & random \\                   
10&	     S-SH 	 & random \\                   
11-14 &	 random 	 & random \\                   
15&	     random 	 & S-SH \\                   
16-21&	 random 	 & random \\                   
22&	     S-SH 	 & random \\                   
23-26&	 random 	 & random \\                   
27&	     random 	 & S-SH \\                   
28-31&	  random 	 & random \\                   
32&	  random 	 & S-SH \\                   
33&	  S-SH 	 & random \\                   
34&	  random 	 & S-SH \\                   
35&	  S-SH 	 & random \\                   
36&	  random 	 & S-SH \\                   
37-40&	  random 	 & random \\                   
41&	  S-SH 	 & random \\                   
42&	  random 	 & random \\                   
43&	  random 	 & S-SH \\                   
44-45&	  random 	 & random \\                   
46&	  S-SH 	 & random \\                   
47&	  random 	 & random \\                   
\hline                   
\end{tabular}                   
\ \ \ \                    
\begin{tabular}{|c|c|c|}                   
\hline                   
Budget & Worst method & Best method \\                   
\hline                   
\hline                   
48-49&	  S-SH 	 & random \\                   
50&	  random 	 & S-SH \\                   
51&	  random 	 & random \\                   
52&	  S-SH 	 & random \\                   
53&	  random 	 & random \\                   
54&	  random 	 & S-SH \\                   
55-56&	  random 	 & random \\                   
57&	  random 	 & S-SH \\                   
58&	  random 	 & random \\                   
59-60&	  random 	 & S-SH \\                   
61-62&	  random 	 & random \\                   
63-64&	  random 	 & S-SH \\                   
65-66&	  random 	 & random \\                   
67&	  S-SH 	 & random \\                   
68-69&	  random 	 & random \\                   
70-72&	  random 	 & S-SH \\                   
73&	  random 	 & random \\                   
\hline                   
\end{tabular}                   
\caption{\label{dependency_numtries}Worst and best method for different budgets, for the Tuned setting (from previous experiments, the Tuned setting is the least favorable to S-SH when the budget is small). 3 independent instances of random compete with S-SH. Under the null assumption, S-SH should appear as often on the left and on the right of this table.                   
The frequency of win on the right hand side table corresponds to a speed-up +87\%. The frequency of win on the left hand side table (budget $<$48) corresponds to a speed-up -51\%, i.e. detrimental results - but the frequency of worst result is smaller than under null assumption ``S-SH equivalent to random''.}                   
\end{table}                   
TODO we have more elsewhere
}
\subsection{Performance as a function of the number of epochs}

\begin{figure}                   
\center                   
\includegraphics[width=.7\linewidth]{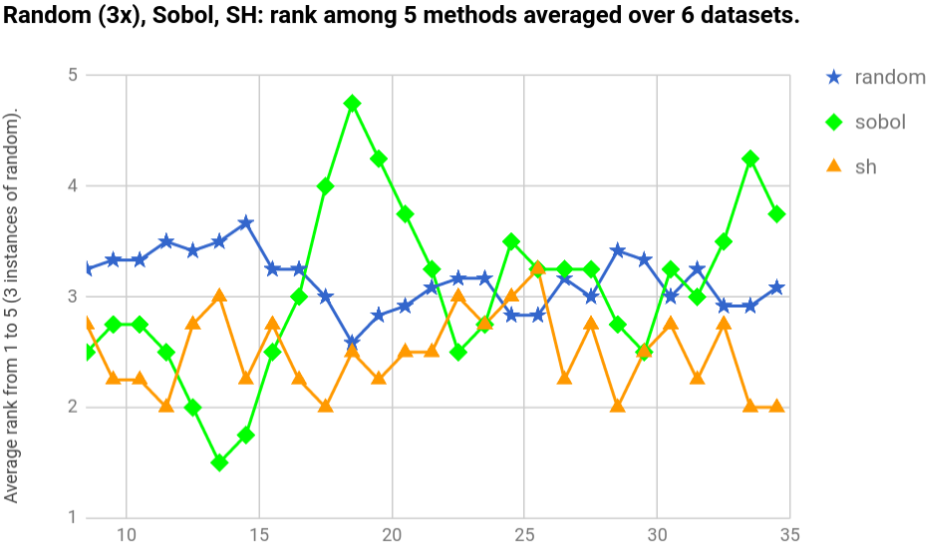}                   
	\caption{\label{impactofmaxepoch}Untuned setting; rank of each method among 5 methods (3 instances of random, plus Sobol and Scrambled-Hammersley), averaged over 6 datasets and 4 successive numbers of epochs (i.e. each point is the average of 24 results and results 3 points apart are independent). This figure corresponds to a budget of 20 vectors of HPs; Scrambled Hammersley outperforms random; Sobol did not provide convincing results. We also tested a budget 10 and did not get any clear difference among methods.}
\end{figure}

We consider the untuned setting, on the same 6 problems as (*) in Fig. \ref{main_table}. We have a number of epochs ranging 
from 7 to 36, and we consider moving averages of the ranks over the 6 datasets and 4 successive numbers of epochs.                   
Fig. \ref{impactofmaxepoch} presents the impact of the number of epochs; S-SH performs best overall.                   

\subsection{Results as an initialization for GP}\label{iter}

\begin{figure}
\center
\includegraphics[width=.32\linewidth]{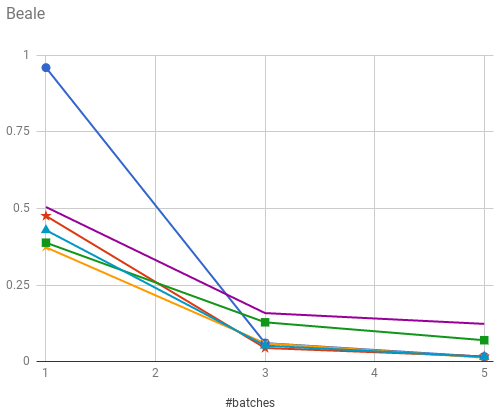}
\includegraphics[width=.32\linewidth]{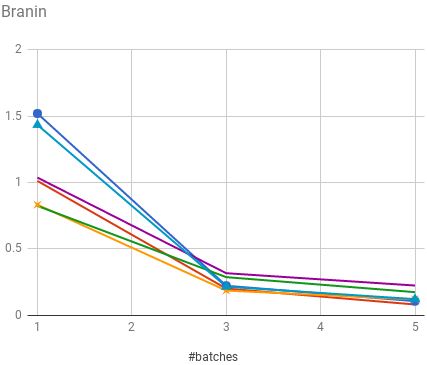}
\includegraphics[width=.32\linewidth]{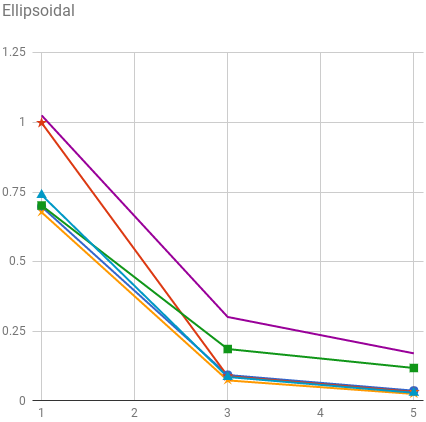}
\includegraphics[width=.32\linewidth]{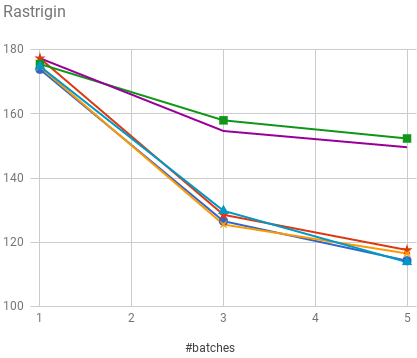}
\includegraphics[width=.32\linewidth]{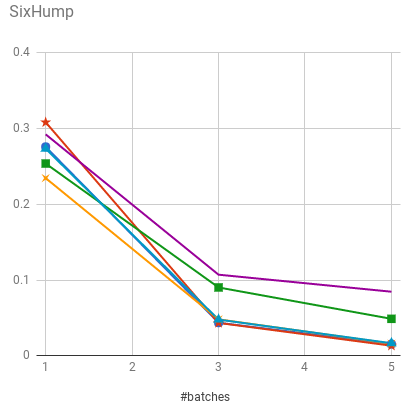}
\includegraphics[width=.32\linewidth]{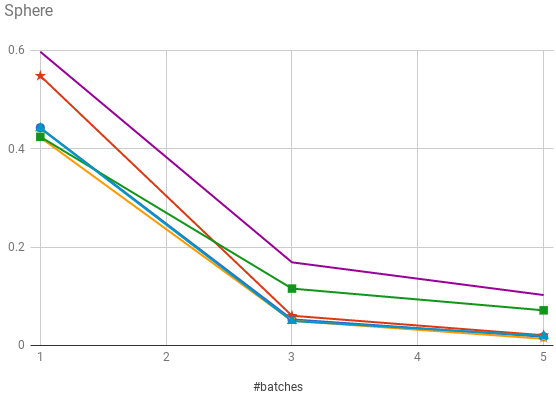}
\includegraphics[width=.32\linewidth]{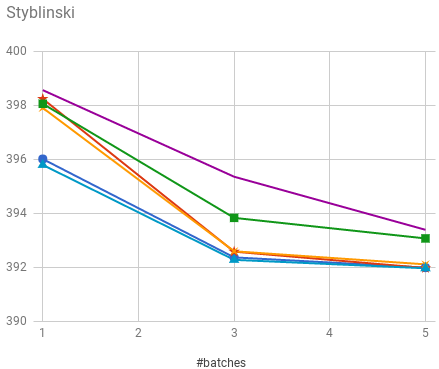}
\includegraphics[width=.2\linewidth]{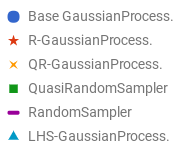}
	\caption{\label{qrgp}Experiments comparing 1, 3 and 5 batches of 64 points for Bayesian optimization (four flavors: base (i.e. pessimistic-fantasizing of unseen values as in Entropy search\cite{fanta}), random, LDS (here Scrambled Halton)) and randomly shifted Scrambled-Halton and random on
	seven test functions. We note that LDS-based Bayesian optimization usually outperformed all other methods.}
\end{figure}

In this section we leave the one-shot setting; we consider several batches, and a Gaussian processes (GP) based Bayesian optimization.
Results in \cite{qrgpclean} already advocated LDS for the initialization of GP, pointing out that this outperforms Latin Hypercube Sampling;
in the present section, we confirm those results in the case of deep learning and show that we also outperform typical pessimistic fantasizing as used in batch 
entropy search.
The pessimistic fantasizing method for GP is based on (i) for the first point of a batch, use optimistic estimates on the value as a criterion
for selecting candidates (ii) for the next points in the same batch, fantasizing the values of the previous points in the current batch
in a pessimistic manner; and apply the same criterion as above, assuming these pessimistic values for previous values of the batch.
For the first batch, this leads to regular patterns as in grids. We keep the same method for further iterations but replace this first
batch by (a) random sampling (b) low-discrepancy sampling.
Results are presented in Fig. \ref{qrgp}. We randomly translate
the optimum in the domain but do not rotate the space of functions - which would destroy the concept of critical variables. We consider
artificial test functions, namely Sphere, Ellipsoid, Branin, Rastrigin, Six-Hump, Styblinski, Beale. 
We work in dimension 12, with 64 vectors of HPs per batch.
We average results over 750, 100 and 50 runs for 1, 3 and 5 batches respectively.

\begin{table}
	\center
	Dimension 12, budget 4 \\
  \includegraphics[width=.68\linewidth]{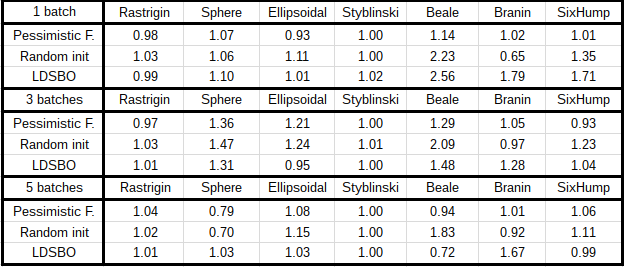}
	Dimension 12, budget 16 \\
  \includegraphics[width=.68\linewidth]{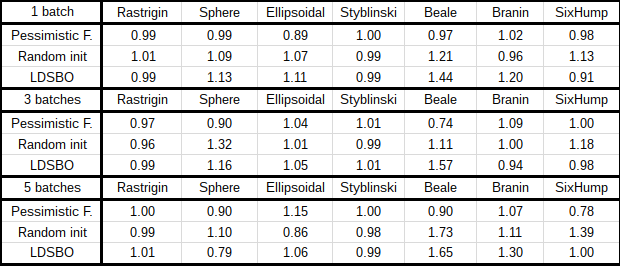}\\
	\caption{\label{gptable16}Experiments with lower budget, in which case LHS (which has excellent discrepancy for low budget) performs well. Ratio ``loss of a method / loss of LHS-BO method'' (i.e. $>1$ means LHS-BO wins) in dimension 12. We see that LHS-BO outperforms LDS-BO for the small budget 4 (i.e. most values are $>1$). For budget 16, pessimistic fantasizing can compete. BO with randomized initialization is always weak compared to LHS-BO.}
\end{table}

\end{document}